\documentclass[conference]{IEEEtran}
\IEEEoverridecommandlockouts
\usepackage[T1]{fontenc}
\usepackage{cite}
\usepackage{amsmath,amssymb,amsfonts}
\usepackage{graphicx}
\usepackage{textcomp}
\usepackage{xcolor}
\usepackage{url}
\def\BibTeX{{\rm B\kern-.05em{\sc i\kern-.025em b}\kern-.08em
    T\kern-.1667em\lower.7ex\hbox{E}\kern-.125emX}}
\newcommand{\artifacturl}{\url{https://anonymous.4open.science/r/TelemetrySuffBench-E635/README.md}}
\begin{document}

\title{TelemetrySuffBench: Is Agent Telemetry Sufficient for Failure-Origin Diagnosis?}

\author{
\IEEEauthorblockN{Yuxuan Zhu, Peng Pu\textsuperscript{*}}
\IEEEauthorblockA{
\textit{School of Data Science and Engineering}\\
\textit{East China Normal University}\\
Shanghai, China\\
Email: 51285903080@stu.ecnu.edu.cn, ppu@cc.ecnu.edu.cn
}
\thanks{\textsuperscript{*}Peng Pu is the corresponding author. Email: ppu@cc.ecnu.edu.cn.}
}

\maketitle

\begin{abstract}
Agent systems increasingly expose execution traces, yet telemetry that reveals a failure may still be inadequate for identifying where that failure originated. We introduce TelemetrySuffBench, a controlled benchmark that separates failure detection, fault-origin localization, and safe abstention under insufficient evidence. The benchmark constructs canonical multi-component traces with delayed-binding faults and renders them as paired coarse views, seven-factor telemetry masks, and exact-equal ambiguous origin pairs. We evaluate five frontier language models using unified protocols, explicit candidate sets, invalid-output accounting, subgroup analyses, and a frozen blind holdout. With full telemetry, origin-step Top-1 accuracy ranges from 33.8\% to 97.2\% across models. Metadata, OpenTelemetry-compatible, and OpenInference-compatible views retain 99.5\% to 100\% detection F1 while limiting origin-step accuracy to at most 0.5\%, exposing a robust detection--localization gap. Factor ablations further show that removing decision content reduces origin-step accuracy to zero for every model, while provenance removal also causes large model-dependent losses. On rich ambiguous inputs that require abstention, evidence gating reduces unsupported unique-origin answers by 12.5 to 48.6 percentage points for three models, whereas two models still answer every case, revealing strong model dependence in safe abstention. Results on the frozen holdout reproduce the central pattern within the same generator family. These findings show that terminal status can support detection, whereas reliable causal attribution requires explicit decision-to-provenance links and abstention safeguards that remain effective across models. The dataset and benchmark implementation are available at \artifacturl.
\end{abstract}


\section{Introduction}
Large language model (LLM) agents increasingly execute tasks through multi-stage workflows that combine model decisions, memory retrieval, tool calls, inter-agent handoffs, and environment observations. These systems produce growing volumes of execution telemetry, and emerging AgentOps practices treat such telemetry as a foundation for monitoring and debugging autonomous behavior \cite{dong2024agentops}. Long-running services and concurrent agent fleets turn per-workflow events into continuous, high-dimensional observability streams spanning models, tools, memory systems, and environments. At this scale, the challenge is not only storing and querying telemetry, but also determining which fields preserve the causal evidence needed for automated diagnosis. Yet observability alone does not establish diagnostic adequacy. A terminal status can reveal that an execution failed, while the event that introduced the failure may lie several causal stages upstream. When a latent decision error propagates through otherwise valid tool calls, a diagnostician can easily select the first visible deviation, a downstream symptom, or the terminal verifier instead of the originating event.

Recent work has made failure attribution in agent systems an explicit evaluation target. Who\&When annotates failure-responsible agents and decisive error steps across logs from 127 multi-agent systems \cite{zhang2025whowhen}; TraceElephant studies attribution from complete and partial multi-agent traces \cite{chen2026traceelephant}; and AgentRx localizes critical failure steps from execution trajectories using constraint-based diagnosis \cite{barke2026agentrx}. These benchmarks and methods establish the importance and difficulty of agent-level and step-level diagnosis. They generally evaluate attribution from a selected trace representation, leaving a prior question unresolved: which parts of that representation actually make the labeled origin identifiable? Without this distinction, a high detection score can mask an absence of origin evidence, and a localization score can conflate causal attribution with recognition of a downstream manifestation.

This distinction is familiar in root-cause analysis for distributed systems, where benchmarks evaluate coarse- and fine-grained localization over metrics, logs, and traces \cite{pham2024rcaeval}, and causal methods connect observed changes to candidate interventions \cite{li2022circa}. Two telemetry standards frame the representation question studied here. OpenTelemetry is a vendor- and tool-agnostic observability framework for generating, collecting, and exporting signals such as traces, metrics, and logs; it also defines a protocol and semantic conventions for representing those signals \cite{opentelemetry2026overview}. OpenInference complements OpenTelemetry with conventions and instrumentation for tracing AI applications, including LLM invocations, retrieval context, and external tool use \cite{openinference2026spec}. Our OpenTelemetry- and OpenInference-compatible views are deterministic projections of the canonical trace onto fields expressible through these respective conventions. Agent executions add semantic dependencies that a schema-level projection may not preserve: a model may select a symbolic reference, a registry may bind it to a concrete object, and a later component may activate that object correctly even when the original selection was wrong. Telemetry sufficiency is therefore task-specific. Evidence adequate for failure detection, fault classification, component localization, and exact origin-step localization need not be the same. A useful evaluation must separate these diagnostic levels and record where an attribution lands along the causal path.

Insufficient origin evidence also creates a selective-prediction problem. Classifiers can trade coverage for lower risk by rejecting uncertain inputs \cite{geifman2017selective}, while unanswerable-question benchmarks test whether language models refrain from producing unsupported answers \cite{rajpurkar2018squad2,madhusudhan2024abstention}. Failure attribution requires a stricter construction: an input is unanswerable when the visible telemetry is compatible with multiple distinct causal origins. Merely hiding fields or instructing a model to abstain does not establish this property. Matched traces with identical model-visible payloads but different injected origins provide a direct test of whether a model respects evidential non-identifiability or supplies an unsupported unique cause.

We introduce \emph{TelemetrySuffBench}, a controlled benchmark for measuring telemetry sufficiency across three linked questions. RQ1 asks whether views that support failure detection also support localization of the injected origin. RQ2 decomposes telemetry into seven semantic factors---identity and event semantics, decision content, provenance mapping, propagation relations, tool and state information, verifier evidence, and terminal status---and measures their contribution through paired masks. RQ3 evaluates safe abstention on exact-equal ambiguous origin pairs and tests whether explicit evidence-gating prompts reduce unsupported attribution. The benchmark represents executions in a versioned canonical trace format, deterministically renders every view from the same underlying run, scores invalid outputs as errors, and reports component, event, causal-stage, coverage, and abstention outcomes. We evaluate five frontier language models under a unified protocol and repeat the central analyses on a frozen seeded holdout.

The results expose a consistent separation between detecting a failure and locating its source. With full telemetry, origin-step Top-1 accuracy ranges from 33.8\% to 97.2\% across models. Metadata, OpenTelemetry-compatible, and OpenInference-compatible views preserve 99.5\% to 100\% detection F1 but limit origin-step accuracy to at most 0.5\%. Removing decision content reduces step localization to zero for every model; removing provenance mapping produces further large, model-dependent losses. On rich ambiguous inputs that require abstention, evidence gating reduces unsupported unique-origin answers by 12.5 to 48.6 percentage points for three models, while two models continue to answer every case. The frozen holdout reproduces the central detection--localization pattern within the same delayed-binding generator family.

This paper contributes a task hierarchy that spans detection, classification, component localization, origin-step localization, and safe abstention, together with causal-stage analysis that distinguishes origins from downstream manifestations. It realizes this formulation in a deterministic benchmark combining paired coarse views, seven-factor masks, and exact-equal ambiguous origin pairs over canonical multi-component traces. A unified five-model evaluation adds explicit candidate sets, invalid-output accounting, operator/domain/source subgroups, prompt interventions, and a frozen blind holdout, yielding reproducible evidence about which telemetry supports causal diagnosis.

\section{Background and Related Work}

\subsection{Failure Attribution in LLM Agent Systems}

Failure attribution has emerged as a distinct problem from end-task evaluation. Who\&When defines attribution at two levels---the responsible agent and the decisive error step---and provides fine-grained annotations over failure logs from 127 multi-agent systems \cite{zhang2025whowhen}. TraceElephant focuses on the evidence exposed to an attribution method: its full-observability benchmark includes agent inputs and execution context that partial, output-only traces omit, and reports gains of up to 76.5\% from using full traces \cite{chen2026traceelephant}. AgentRx contributes 115 manually annotated failed trajectories across structured API workflows, incident management, and open-ended web/file tasks, together with a cross-domain taxonomy and a constraint-based diagnostic method for locating critical steps \cite{barke2026agentrx}. Collectively, these benchmarks establish responsible-agent and failure-step localization as difficult, operationally relevant tasks.

Controlled-execution methods use interventions to improve attribution: AgenTracer generates training trajectories through counterfactual replay and fault injection, while REFLECT tests candidate causes through diagnosis-specific patches and replay \cite{zhang2025agentracer,lin2026reflect}. TelemetrySuffBench addresses the complementary observational question: does the visible trace distinguish the injected origin from its activation, first visible deviation, downstream symptom, and terminal manifestation? Its paired views hold the execution and label fixed while varying only the evidence presented to the model.

\subsection{Telemetry Standards and Root-Cause Analysis}

OpenTelemetry provides vendor-neutral APIs, protocols, and semantic conventions for generating and transporting traces, metrics, and logs \cite{opentelemetry2026overview}. OpenInference extends this foundation with conventions and instrumentation for AI application traces, including LLM invocations, retrieval context, and external tool use \cite{openinference2026spec}. AgentOps broadens the observability scope across the agent lifecycle by identifying artifacts and data needed for monitoring, logging, and analytics \cite{dong2024agentops}. These efforts standardize how execution evidence is captured and exchanged. They do not, by themselves, determine whether a particular projection contains the semantic links required for a given diagnostic target.

Root-cause analysis for microservices provides the closest systems precedent. RCAEval assembles 735 failure cases from three systems and evaluates 15 reproducible baselines at coarse and fine localization granularities \cite{pham2024rcaeval}. CIRCA formulates root-cause analysis as intervention recognition over a causal Bayesian network \cite{li2022circa}. This literature asks how to improve localization from available telemetry. TelemetrySuffBench treats the telemetry representation as the independent variable and measures which diagnostic levels remain answerable after controlled removal of identity, decision, provenance, relation, state, verification, or terminal signals.

\subsection{Abstention and Evidential Identifiability}

Selective classification formalizes abstention as a risk--coverage trade-off, allowing a predictor to reject inputs to satisfy a target risk level \cite{geifman2017selective}. Language-understanding benchmarks operationalize a related requirement through unanswerable questions. SQuAD~2.0 adds more than 50,000 adversarially written unanswerable questions that resemble answerable ones and requires systems to determine when a passage supports no answer \cite{rajpurkar2018squad2}. Abstain-QA evaluates black-box language models across answerable and unanswerable questions and compares prompting strategies for improving abstention \cite{madhusudhan2024abstention}. AbstentionBench expands the scope to 20 datasets covering unknown answers, underspecification, false premises, subjectivity, and outdated information, finding persistent abstention failures across 20 frontier models \cite{kirichenko2025abstentionbench}.

Failure-origin attribution introduces a different source of unanswerability. The answer can be absent not because the model lacks world knowledge or the request is linguistically incomplete, but because the same visible execution is compatible with multiple distinct injected origins. TelemetrySuffBench constructs this condition directly: each ambiguous pair has exact-equal model-visible payloads and different gold origins. Any unique-origin answer is therefore unsupported by the observation, even if it coincides with one member's hidden label. This construction connects selective prediction to causal diagnosis and permits separate measurement of unsafe answers on ambiguous traces and unnecessary abstention on answerable traces.

\section{Methodology}

\subsection{Task Formulation}

We represent an execution trace as $\tau=(E,G,X,Y)$. Here, $E=(e_1,\ldots,e_n)$ is an ordered event sequence, $G$ contains parent and dependency edges, $X$ contains the event attributes available for rendering, and $Y$ contains private evaluation labels. For a telemetry configuration $m$, a deterministic renderer $g_m$ produces the model-visible view
\begin{equation}
x_m(\tau)=g_m(E,G,X).
\end{equation}
The renderer never reads $Y$. The gold diagnosis for a fault trace is $y=(f,c,o)$, where $f$ is the fault type, $c$ is the component that introduced the fault, and $o$ is its origin event. Clean traces have no fault type or origin.

The evaluation separates four diagnostic decisions. \emph{Detection} predicts whether a fault occurred. \emph{Classification} selects a type from the closed taxonomy. \emph{Localization} identifies the origin component and event. \emph{Answerability} determines whether the visible evidence supports one unique origin. Detection and classification can remain possible when localization is not; RQ3 supplies constructive answerability labels, while RQ1 and RQ2 measure diagnostic performance under controlled changes to the visible evidence.

\begin{figure*}[t]
\centering
\includegraphics[width=0.98\textwidth]{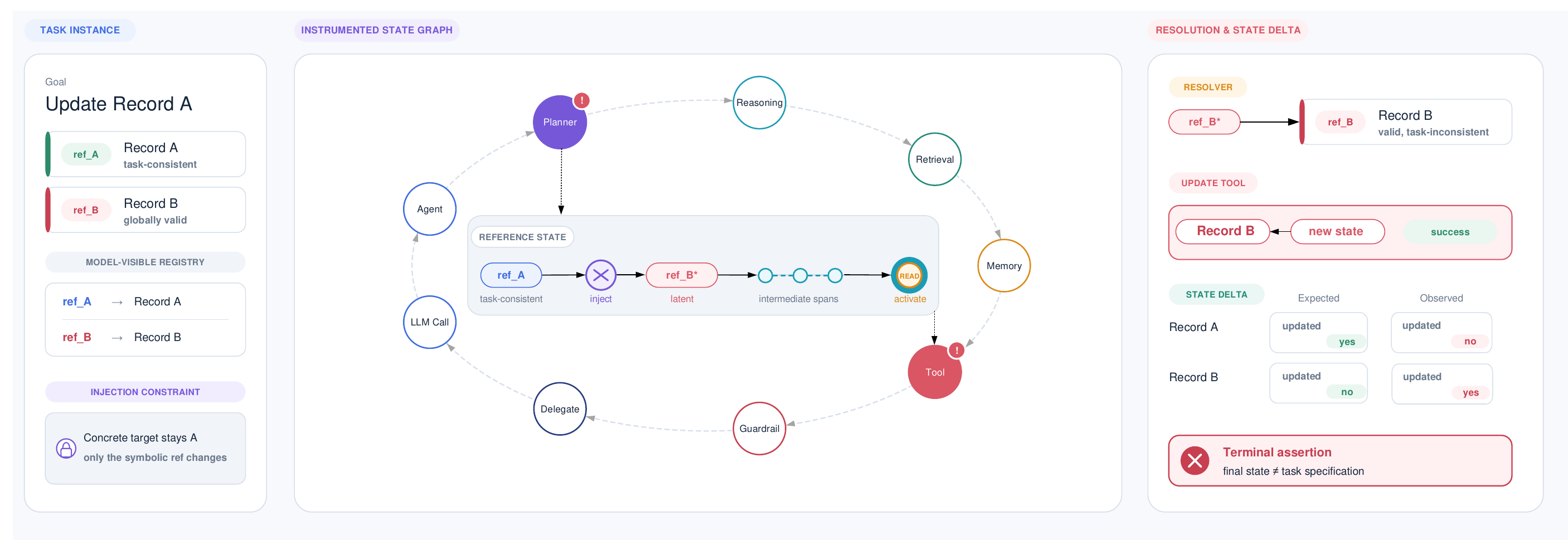}
\caption{Controlled construction of a delayed-binding fault. The task requires Record A, while a globally valid but task-inconsistent symbolic reference is injected at the origin component. After intermediate spans, activation resolves the latent reference to Record B; the update succeeds on B, and the terminal assertion detects the state mismatch.}
\label{fig:construction_overview}
\end{figure*}

\subsection{Dataset Construction}

Figure~\ref{fig:construction_overview} summarizes the controlled fault construction. TelemetrySuffBench is generated deterministically from instrumented, stateful workflows. Each instance asks an agent pipeline to update one designated record and verify the resulting state. The generator instantiates three domains---tickets, documents, and orders---and four cyclic workflow topologies over nine components: agent, planning, reasoning, retrieval, memory, tool call, guard rail, delegation, and LLM call. Each execution emits spans through public AgentTelemetry instrumentation at a fixed source revision \cite{balusu2026agenttelemetry}, followed by a delayed reference activation, a resolver, a state-update tool, and a terminal verifier.

Every task defines a correct record and an alternate record, both of which have valid entries in a visible reference registry. A clean trace carries the task-consistent symbolic reference. To construct a fault, the generator injects a globally valid but task-inconsistent reference at a selected origin component while leaving the immediate concrete target unchanged. One of six operators determines the reference relation: alias binding, version binding, namespace handoff, schema binding, evidence binding, or retry-token binding. After intervening workflow events, an activation consumes the latent reference, the resolver binds it to the alternate record, the update succeeds on that record, and the verifier detects the final-state mismatch. Version and retry-token faults are labeled \texttt{stale\_reference\_state}; the other four operators are labeled \texttt{wrong\_reference\_binding}.

The main corpus contains 96 clean controls and 108 matched two-origin fault groups. The two fault traces in each group share the domain, task variant, topology, operator, registry, resolver behavior, state-update symptom, and terminal outcome, but inject the faulty reference at different components and events. This produces 216 fault traces, with 24 origins at each of the nine components, including 144 \texttt{wrong\_reference\_binding} and 72 \texttt{stale\_reference\_state} cases. Together with the clean controls, the corpus contains 312 traces.

Programmatic hooks record the injected origin, reference activation, first visible wrong-target event, state-update symptom, and terminal failure. These records define the causal-stage labels \emph{origin}, \emph{activation}, \emph{first visible deviation}, \emph{symptom}, and \emph{terminal}. The labels come from the injection hook and deterministic final-state assertion and are stored outside the model-visible trace.

\subsection{Canonical Trace Representation}

Every raw execution is converted into a versioned canonical trace before any view is rendered. An event records its identifier and step index; actor identity, role, display name, and component type; event type; parent and dependency identifiers; input and output; tool name, arguments, and result; retrieval and memory provenance; status and exception; and logical time. Validation requires unique event identifiers, valid graph references, and a visible origin event for every fault trace.

Evaluation labels are stored separately from event attributes. In addition to the fault type and exact origin, they include activation, first-visible-deviation, symptom, terminal, causal-witness, and propagation-edge annotations. Source pointers, injection records, labels, and dataset metadata are excluded wherever the factor renderer defines them as private. Candidate component and event sets are constructed from the corpus and the rendered trace, respectively, and are supplied explicitly to prevent invented labels.

\subsection{RQ1: Paired Coarse Views}

RQ1 renders each canonical execution into six deterministic views, keeping the underlying run and private label fixed. \emph{Full} contains the canonical trace except its private \texttt{labels} and \texttt{metadata} objects. \emph{Content-redacted} retains event identities, types, statuses, and relations while recursively replacing present input, output, tool, retrieval, memory, and exception values with \texttt{<REDACTED>}. \emph{Metadata} retains event identity and type, status, tool name and arguments, and parent/dependency relations, while omitting tool results and exceptions. \emph{Structural} retains event identifiers, indices, types, statuses, and parent/dependency relations. The \emph{OpenTelemetry-compatible} and \emph{OpenInference-compatible} views retain only canonical identity, tool, observation, status, and relation fields that have direct generic mappings to the corresponding conventions \cite{opentelemetry2026overview,openinference2026spec}. For the canonical fields available in this corpus, the two compatibility renderers use the same conservative generic field set. Thus every within-trace comparison changes only the rendered evidence. Compatibility is defined field-wise by the renderer and does not add information absent from the canonical execution.

\subsection{RQ2: Semantic-Factor Ablations}

RQ2 replaces format-level views with the seven semantic factors in Table~\ref{tab:factors}. The manifest assigns every renderable leaf field to one factor. Disabled fields are replaced by \texttt{<REDACTED>}; event identifiers, step indices, and the candidate event list remain visible under every mask. Labels and injection provenance are never rendered, and any unregistered field causes rendering to fail closed.

\begin{table*}[t]
\caption{Semantic telemetry factors used in RQ2.}
\label{tab:factors}
\centering
\footnotesize
\begin{tabular}{c p{0.25\textwidth} p{0.63\textwidth}}
\hline
\textbf{Factor} & \textbf{Semantic group} & \textbf{Canonical fields} \\
\hline
I & Identity and event semantics & Actor identity and role, display name, component and event type, and tool name \\
D & Decision and latent-reference content & Selected symbolic reference and binding kind \\
P & Registry and provenance mappings & Reference registry map and retrieved provenance documents \\
R & Propagation and handoff relations & Parent and dependency edges and memory source-event links \\
S & Tool inputs and state transitions & Inputs, general outputs, tool arguments/results, retrieval queries, and memory operations, keys, and versions \\
V & Observation and verifier evidence & Expected and actual target values \\
T & Terminal failure status & Event status and exception text \\
\hline
\end{tabular}
\end{table*}

Seven binary factors define $2^7=128$ possible masks. Exhaustively evaluating this lattice for every model would multiply model calls while adding many redundant comparisons, whereas choosing arbitrary subsets after inspecting model outputs would introduce mask-selection overfitting. We therefore fix an 11-mask cross-model panel around two interpretable contrast families. Full and the seven leave-one-factor-out masks---Full-I, Full-D, Full-P, Full-R, Full-S, Full-V, and Full-T---estimate the effect of removing one semantic group. Three targeted subsets test compact, theory-driven combinations: I+D+P+V retains identity, decision, provenance, and verifier evidence; D+P+V removes identity from that localization-oriented core; and S+V+T isolates downstream tool/state, verifier, and terminal evidence. Every trace is evaluated under the same 11 masks, and the panel is held fixed across models. This paired design measures removal effects and tests whether interpretable compact combinations preserve detection, classification, or localization.

\noindent\textbf{Full-D matched-pair proposition.} Let $z_F(\tau)$ be the visible fields assigned to factor $F$. If a matched pair has equal $z_F$ for every $F\in\{I,P,R,S,V,T\}$ and equal always-visible fields and candidate sets, then its complete Full-D payloads are equal. The proof follows directly from the deterministic renderer, which preserves those equal fields and maps every D-owned leaf to the same token. The generator enforces these premises, and canonical serialization verifies the conclusion for all 108 groups. This is not a claim about arbitrary cross-group traces, whose retained factors may differ.

\subsection{RQ3: Exact Observational Ambiguity}

RQ3 constructs unanswerability from observational equivalence. Let $P_m(\tau)$ denote the complete canonicalized model payload under mask $m$, including the instruction, candidate component and event lists, rendered telemetry, and output schema. A matched pair $(\tau_a,\tau_b)$ is ambiguous under $m$ only if
\begin{equation}
P_m(\tau_a)=P_m(\tau_b), \qquad \omega_a\neq\omega_b,
\label{eq:ambiguity}
\end{equation}
where $\omega=(c,o)$ is the injected origin. Eligibility further requires exactly two traces in the group and the same terminal symptom. Equality in \eqref{eq:ambiguity} is byte-stable equality after canonical JSON serialization of the complete payload.

Each eligible group yields four conditions. The compact-answerable mask I+D+P+V and the richer answerable mask I+D+P+R+V+T must produce different payloads for the two origins. Among all exact-equal masks, the sparse-ambiguous condition selects the candidate with the fewest factors and lowest payload richness, while the rich-ambiguous condition selects the candidate with the most factors and highest payload richness; the rich mask must contain strictly more factors than the sparse mask. Answerable rows are labeled \texttt{ANSWERABLE} with the exact component and event. Both members of an ambiguous pair are labeled \texttt{INSUFFICIENT\_EVIDENCE} with null origins. A unique answer on an ambiguous payload is unsupported even when it happens to match one member's private label.

The primary P0 instruction states the abstention rule directly. Three prompt variants change only the decision policy: P1 is a minimal paraphrase, P2 requires explicit evidence connecting and distinguishing one origin, and P3 asks the model to eliminate alternative origin hypotheses before answering. P0 was fixed first; P1--P3 were jointly frozen before their model calls. The blind holdout retains P0 and P3 as the direct and strongest contrastive policies.

\subsection{Output Protocol, Metrics, and Uncertainty}
\label{sec:output_protocol}

All protocols request JSON and provide closed candidate sets. RQ2 requires exactly \texttt{fault\_present}, \texttt{fault\_type}, \texttt{answerability}, \texttt{origin\_component}, and \texttt{origin\_event\_id}; RQ3 uses the final three fields. An \texttt{ANSWERABLE} fault requires a candidate component and visible candidate event, whereas \texttt{INSUFFICIENT\_EVIDENCE} requires both origin fields to be null. Malformed, inconsistent, out-of-candidate, and out-of-taxonomy outputs remain in the evaluation denominator and receive no correctness credit.

RQ1 and RQ2 report detection F1 over all traces and fault-type Macro-F1 over gold fault traces under the two-class taxonomy. For $N_F$ fault traces, component and origin-step accuracy are
\begin{align}
\mathrm{Acc}_{\mathrm{comp}}&=\frac{1}{N_F}\sum_{i\in F}\mathbf{1}[\hat c_i=c_i],\\
\mathrm{Acc}_{\mathrm{step}}&=\frac{1}{N_F}\sum_{i\in F}\mathbf{1}[\hat o_i=o_i].
\end{align}
Invalid outputs, abstentions, and missing origins contribute zero. Coverage is the proportion of fault rows on which the model returns a valid unique-origin answer. Causal-stage analysis assigns each selected event to origin, activation, first visible deviation, symptom, terminal, other, abstain, or invalid. When stage labels coincide, assignment follows that listed order.

Four deterministic stage-collapse references contextualize localization accuracy. Each reference assigns every fault to one annotated causal stage and returns that event together with its owning component. The activation reference selects the resolver event that first consumes the latent reference. The first-visible reference selects the first event whose resolved target conflicts with the task-consistent registry expectation. The symptom reference selects the wrong-target state update, and the terminal reference selects the failed verifier. These mask-invariant references quantify the result of systematically collapsing causal attribution onto one downstream stage. They are scored on the 216 RQ1 fault traces and the 144 RQ2 confirmation faults.

For RQ3, let $A$ and $U$ be the answerable and ambiguous rows. The false-answer rate (FAR) and unnecessary-abstention rate (UAR) are
\begin{align}
\mathrm{FAR}&=\frac{1}{|U|}\sum_{i\in U}\mathbf{1}[\hat a_i=\texttt{ANSWERABLE}],\\
\mathrm{UAR}&=\frac{1}{|A|}\sum_{i\in A}\mathbf{1}[\hat a_i=\texttt{INSUFFICIENT\_EVIDENCE}],
\end{align}
where the indicators require a valid output. FAR therefore measures unsupported unique-origin answers, and 100\% is the worst value. We also report valid-answer coverage, joint component-and-event localization on answerable rows, invalid-output rate, and safe accuracy: correct origins on $A$ plus correct abstentions on $U$, divided by all rows.

Uncertainty analyses use 10,000 percentile bootstrap resamples with seed 20260729. The cluster unit is the matched fault group; clean traces form singleton clusters for RQ2. Joint resampling preserves all compared masks, models, conditions, and prompts within a sampled cluster.

\subsection{Frozen Splits and Blind Holdout}

The primary controlled corpus contains 312 traces. The discovery split has 96 traces (72 fault and 24 clean), and the confirmation split has 216 traces (144 fault and 72 clean). The splits have no trace overlap, never divide a matched fault group, and each covers all three domains, six delayed operators, and nine origin components. Factor definitions, masks, parsers, metrics, and RQ3 group construction were frozen through versioned manifests and content hashes.

A separate blind holdout was generated with seed 20260730 and frozen before model execution. It contains 216 traces: 144 faults arranged into 72 matched groups and 72 clean traces. The holdout uses new task instances from the same delayed-binding generator family and applies the same 11 RQ2 masks, the same four RQ3 conditions, and the P0/P3 prompt policies. It tests sample generalization within the controlled mechanism family while preserving the original output validation and scoring rules.

\subsection{Mechanism and Public-Trace Validation}

An independent evidence-gate pilot tests whether the Full-D localization collapse extends beyond delayed reference binding. The pilot contains five base cases. Each case has five fault variants that differ only in which candidate component accepts unreconciled evidence, plus one clean trace, for 25 fault and five clean traces. Within every base case, the five fault payloads have distinct Full hashes and one identical Full-D hash. GPT-5.6 Sol evaluates Full and Full-D. The confirmatory mechanism-family analysis admits datasets only when they provide injected causal origins, a validated seven-factor crosswalk, matched Full/Full-D equivalence groups, and clean/fault controls.

All five models also evaluate Full and Content-redacted views of 218 scorable TraceElephant traces spanning Captain-Agent, Magentic-One, and SWE-Agent \cite{chen2026traceelephant}. Two traces with nonnumeric decisive-step labels are excluded. Canonicalization uses stable agent names, preserves responses and tool logs, and applies a fixed 3,000-character middle truncation to each request. This failure-only corpus supports localization comparisons.

\section{Experimental Setup}

\subsection{Model Panel}

We evaluated GPT-5.5, GPT-5.6 Sol, DeepSeek-V4-Pro, Claude Opus 5, and Qwen 3.7 Max. Returned model identities and documented aliases were checked against the requested model before scoring. GLM-5.2 was excluded because its returned identity was inconsistent across requests.

\subsection{Evaluation Workloads}

RQ1 evaluates 312 traces under six views, yielding 1,872 logical cases per model. RQ2 combines the 216-trace confirmation split with the fixed 11-mask panel, yielding 2,376 cases per model. RQ3 evaluates 72 matched groups under four telemetry conditions and four prompt policies, yielding 2,304 logical rows per model. The blind holdout adds 2,376 RQ2 cases and 864 RQ3 P0/P3 cases per model. A repeatability study uses 24 matched groups, four RQ2 masks, both rich RQ3 conditions, and the P0/P3 prompts; three repeats yield 1,008 cases per model. The two validations add 2,240 fixed-evaluator requests.

\subsection{Reproducibility and Artifact Availability}

The dataset, benchmark implementation, and evaluation scripts are available at \artifacturl. Reproducing the complete experiment costs approximately US\$2,400 in model API charges.

\section{Results}

\subsection{RQ1: Does Failure-Detecting Telemetry Support Origin Localization?}

\begin{figure*}[t]
\centering
\includegraphics[width=\textwidth]{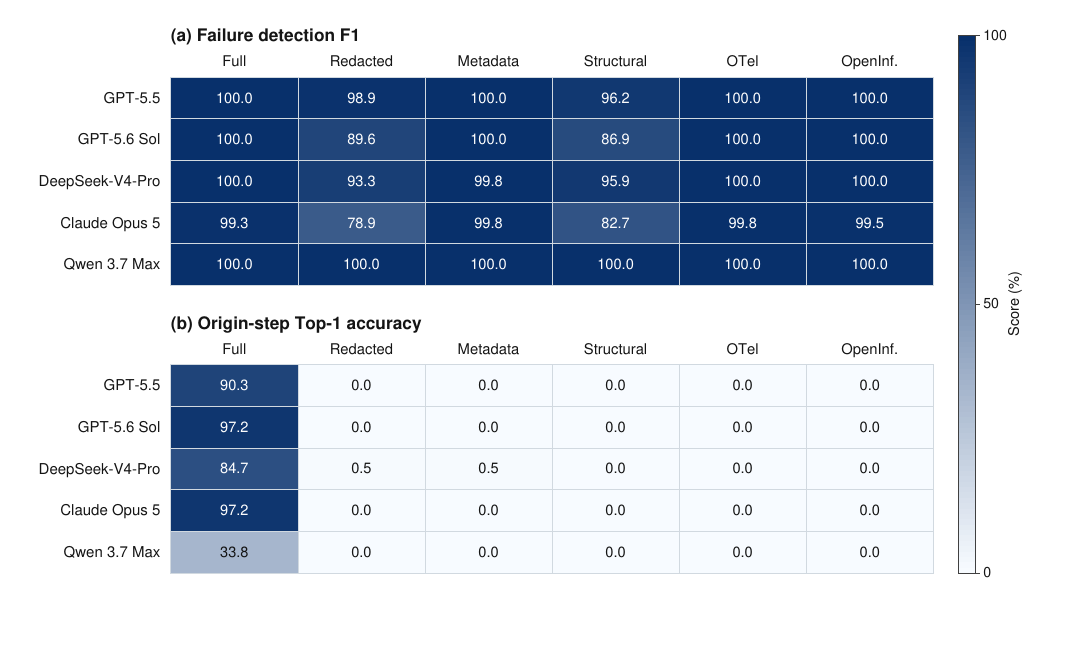}
\caption{RQ1 failure detection and exact origin-step localization across five models and six telemetry views. Each cell is a percentage over the same 312 traces, including 216 fault traces. Restricted views preserve high detection scores while removing nearly all exact-origin accuracy.}
\label{fig:rq1_gap}
\end{figure*}

RQ1 tests whether a telemetry view that exposes a failure also identifies the event that introduced it. Figure~\ref{fig:rq1_gap} reports detection F1 and origin-step Top-1 accuracy for every model--view pair, and Table~\ref{tab:rq1_full} gives the full-view classification and localization results. The two metrics separate sharply once decision and provenance content is removed.

\begin{table}[t]
\caption{RQ1 full-view results (\%). Type F1 is fault-type Macro-F1; stage refs. denotes each of activation, first-visible, symptom, and terminal.}
\label{tab:rq1_full}
\centering
\footnotesize
\begin{tabular}{lrrrr}
\hline
\textbf{Model} & \textbf{Det. F1} & \textbf{Type F1} & \textbf{Comp.} & \textbf{Step} \\
\hline
GPT-5.5 & 100.0 & 79.9 & 91.7 & 90.3 \\
GPT-5.6 Sol & 100.0 & 76.0 & 97.7 & 97.2 \\
DeepSeek-V4-Pro & 100.0 & 55.3 & 85.6 & 84.7 \\
Claude Opus 5 & 99.3 & 77.7 & 97.2 & 97.2 \\
Qwen 3.7 Max & 100.0 & 39.8 & 45.4 & 33.8 \\
Stage-collapse refs. (each) & -- & -- & 11.1 & 0.0 \\
\hline
\end{tabular}
\end{table}

\subsubsection{Full Telemetry.}
All five models detect failures almost perfectly from the Full view, with detection F1 between 99.3\% and 100.0\%. Exact localization is more demanding: component accuracy ranges from 45.4\% to 97.7\%, and origin-step Top-1 ranges from 33.8\% to 97.2\%. GPT-5.6 Sol and Claude Opus 5 both reach 97.2\% step accuracy, followed by GPT-5.5 at 90.3\% and DeepSeek-V4-Pro at 84.7\%. Qwen 3.7 Max reaches 33.8\%. Fault-type Macro-F1 spans 39.8\%--79.9\% despite near-perfect detection. The diagnostic levels therefore remain distinct even when the complete trace is visible.

Each deterministic stage-collapse reference---activation, first-visible deviation, state-update symptom, or terminal verifier---yields 0.0\% origin-step and 11.1\% component accuracy. Activation and first-visible deviation coincide here, and the nine origins are component-balanced. Thus selecting a salient downstream event does not recover the injected origin.

\subsubsection{Restricted Views.}
The Metadata, OpenTelemetry-compatible, and OpenInference-compatible views yield detection F1 between 99.5\% and 100.0\% across all models. Their component accuracies cluster between 9.7\% and 11.1\%, near the balanced one-of-nine component baseline, while origin-step accuracy never exceeds 0.5\%. Structural telemetry also gives 82.7\%--100.0\% detection F1 but zero step accuracy for every model. Content redaction lowers detection more unevenly, to 78.9\%--100.0\%, yet its step accuracy remains at or below 0.5\%. Terminal status and downstream observations are therefore sufficient to recognize most failures, whereas exact localization depends on trace content that connects a decision to its provenance and later effects.

The causal-stage results show where the restricted-view predictions move. Across the 1,080 fault evaluations for each view, 97.9\% of Metadata predictions, 99.0\% of OpenTelemetry-compatible predictions, and 98.7\% of OpenInference-compatible predictions land on the downstream symptom. Structural outputs contain no origin selections: 47.7\% abstain, 38.9\% select the terminal manifestation, and 12.7\% are invalid. The models thus respond systematically to the strongest remaining evidence, but that evidence describes manifestation rather than origin.

\subsubsection{Subgroup Analysis.}
Under Full telemetry, the operator-specific step-accuracy ranges are 86.1\%--100.0\% for GPT-5.5, 91.7\%--100.0\% for GPT-5.6 Sol, 69.4\%--97.2\% for DeepSeek-V4-Pro, and 91.7\%--100.0\% for Claude Opus 5. Their domain- and origin-component-specific results show the same high-localization regime. Qwen 3.7 Max is less stable, ranging from 2.8\% to 66.7\% across operators, 20.8\% to 44.4\% across domains, and 0.0\% to 66.7\% across origin components. The aggregate separation is not produced by one operator or domain. It combines a view-level loss of origin evidence with model-specific ability to use the Full trace.

\textbf{RQ1 answer.} In this benchmark, failure detection does not establish fault-origin identifiability. Full telemetry supports strong localization for four models, while the restricted views retain enough downstream evidence for near-perfect detection but almost no information that the evaluated models can use to recover the injected origin.

\subsubsection{Public-Trace Stress Test.}
Across all five models, Full yields 9.6\%--24.3\% decisive-event Top-1 and 16.1\%--41.7\% component accuracy on TraceElephant, while Content-redacted yields zero for both metrics. All ten overall paired bootstrap intervals exclude zero, with exact McNemar $p\leq9.54\times10^{-7}$. The direction also holds across Captain-Agent, Magentic-One, and SWE-Agent for every model and metric, extending the RQ1 content-sensitivity pattern to public natural-failure traces.

\subsection{RQ2: Which Telemetry Factors Support Origin Localization?}

\begin{figure*}[t]
\centering
\includegraphics[width=\textwidth]{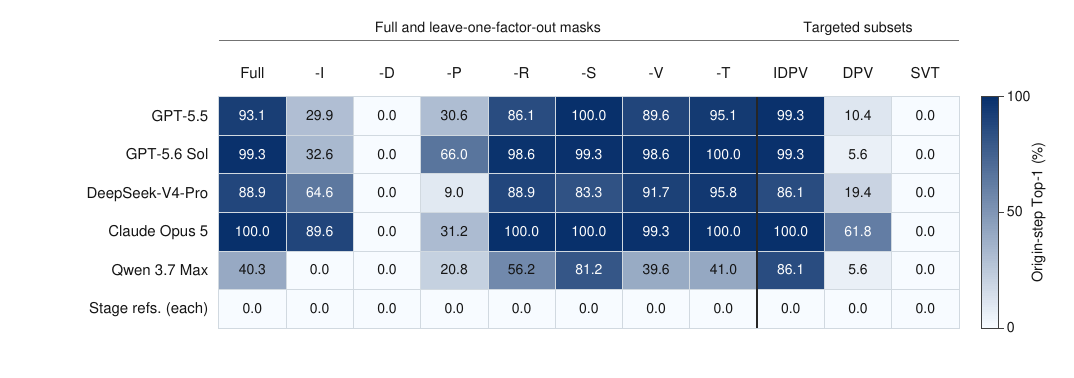}
\caption{RQ2 origin-step Top-1 across the 11 masks. Columns $-X$ remove factor $X$; IDPV, DPV, and SVT are targeted subsets. The final row denotes each stage-collapse reference. Full-D, SVT, and all references yield zero exact-origin accuracy.}
\label{fig:rq2_masks}
\end{figure*}

RQ2 isolates the semantic factors responsible for the localization behavior observed in RQ1. Figure~\ref{fig:rq2_masks} reports origin-step Top-1 accuracy across the fixed 11-mask panel. Full-view accuracy ranges from 40.3\% to 100.0\%, but the paired ablations reveal a consistent structure beneath this model-level variation. Table~\ref{tab:rq2_effects} reports the three principal paired contrasts with 95\% cluster-bootstrap intervals.

\begin{table*}[t]
\caption{Key RQ2 paired effects in percentage points with 95\% cluster-bootstrap intervals. Positive values are localization losses: $\Delta_D=\text{Full}-\text{Full-D}$, $\Delta_P=\text{Full}-\text{Full-P}$, and $\Delta_I^{\mathrm{core}}=\text{IDPV}-\text{DPV}$.}
\label{tab:rq2_effects}
\centering
\footnotesize
\begin{tabular}{lcccc}
\hline
\textbf{Model} & $\boldsymbol{\Delta_D}$ & $\boldsymbol{\Delta_P}$ & $\boldsymbol{\Delta_I^{\mathrm{core}}}$ & \textbf{SVT Step} \\
\hline
GPT-5.5 & 93.1 [87.7, 97.3] & 62.5 [53.5, 71.4] & 88.9 [83.3, 93.8] & 0.0 \\
GPT-5.6 Sol & 99.3 [97.8, 100.0] & 33.3 [24.3, 42.4] & 93.8 [89.6, 97.3] & 0.0 \\
DeepSeek-V4-Pro & 88.9 [83.9, 93.5] & 79.9 [72.6, 86.6] & 66.7 [58.8, 74.3] & 0.0 \\
Claude Opus 5 & 100.0 [100.0, 100.0] & 68.8 [60.0, 77.1] & 38.2 [30.1, 46.6] & 0.0 \\
Qwen 3.7 Max & 40.3 [32.0, 48.6] & 19.4 [12.5, 26.6] & 80.6 [75.0, 86.2] & 0.0 \\
\hline
\end{tabular}
\end{table*}

\subsubsection{Decision Content.}
Removing D reduces origin-step accuracy to zero for all five models and component accuracy to 11.1\%, the balanced one-of-nine baseline. Fault-type Macro-F1 remains between 40.0\% and 88.1\% under Full-D, so the models can still classify the visible failure while losing the information needed to identify its origin. The paired step loss ranges from 40.3 to 100.0 percentage points, and every bootstrap interval excludes zero. Across all 720 model--fault evaluations under Full-D, 98.1\% of predictions select the activation event. Decision removal therefore shifts attribution to the point where the latent reference is resolved rather than the earlier event that selected it.

This result holds at every controlled granularity. Full-D step accuracy is zero for every operator, domain, and origin-component subgroup for all five models. The zero is thus a property of the decision-content ablation, not an average produced by a small set of difficult sources.

\subsubsection{Provenance and Identity.}
Removing P lowers step accuracy to 9.0\%--66.0\%, corresponding to paired losses of 19.4--79.9 percentage points. All five bootstrap intervals exclude zero, while their different magnitudes show that models vary in how effectively they recover or compensate for a missing registry and provenance mapping. The causal-stage distribution moves in the same direction: 66.7\% of Full-P predictions select the activation event and only 31.5\% retain the injected origin.

The targeted subsets separate identity from the rest of the localization core. I+D+P+V achieves 86.1\%--100.0\% step accuracy across models, whereas D+P+V reaches only 5.6\%--61.8\%. Restoring identity yields gains of 38.2--93.8 percentage points, with every interval excluding zero. Identity is therefore most informative in conjunction with decision and provenance content: it names the actor or event to which those semantic links attach.

\subsubsection{Downstream Evidence.}
The S+V+T subset contains tool/state transitions, verifier observations, and terminal status but omits identity, decision, provenance, and propagation relations. Its origin-step accuracy is zero for every model. By contrast, removing S from Full produces 81.2\%--100.0\% step accuracy and matches or exceeds Full for four models. Removing R, V, or T also produces broad model-specific ranges---56.2\%--100.0\%, 39.6\%--99.3\%, and 41.0\%--100.0\%, respectively---and sometimes improves localization. These factors describe propagation and manifestation, but they do not substitute for the decision-to-provenance link. Additional downstream detail can also draw a model toward a later event.

\subsubsection{Holdout and Repeatability.}
The holdout preserves the central contrasts: Full accuracy changes by at most 4.9 points, Full-D and S+V+T remain at zero, and I+D+P+V retains 76.4\%--99.3\% accuracy. Across three repeats on 24 matched groups, Full-D stays at 0\%--2.1\%, Full-S at 75.0\%--100.0\%, and Full-P at 12.5\%--66.7\%. The ordering is more stable than individual predictions.

\textbf{RQ2 answer.} Exact origin localization requires telemetry that exposes the faulty decision and connects it through identity and provenance to a concrete event. Decision content is indispensable across models. Provenance and identity provide large, consistently positive contributions. Downstream state, verification, and terminal evidence cannot recover the origin on their own.

\subsection{Second-Mechanism Replication}

The matched evidence-gate pilot reproduces the central RQ2 result under a second controlled fault mechanism. On its 25 fault traces, GPT-5.6 Sol reaches 92.0\% origin-component and 92.0\% origin-event accuracy with Full. Both localization measures fall to 0\% under Full-D, while detection accuracy remains 96.7\%. In 23 of 25 paired fault cases Full alone identifies the correct component and event; Full-D never identifies either. Full-D predictions predominantly select the common downstream symptom event. Thus two distinct controlled mechanisms---delayed reference binding and unreconciled-evidence acceptance---show the same separation: downstream telemetry preserves fault visibility, while the faulty decision is required to recover the origin.

\subsection{RQ3: Do Models Abstain When the Origin Is Not Identifiable?}

RQ3 tests whether models withhold a unique origin when two distinct origins produce exactly the same model-visible payload. Each condition contains 144 logical rows per model and prompt. Table~\ref{tab:rq3_p0_p3} reports Rich-Ambiguous FAR, Rich-Answerable UAR, and matched P0--P3 contrasts with 95\% intervals from 10,000 matched-group bootstrap resamples.

\begin{table*}[t]
\caption{RQ3 Rich-Ambiguous FAR and Rich-Answerable UAR under P0 and P3. FAR contrasts are P3 minus P0 in percentage points with 95\% matched-group bootstrap intervals. Negative contrasts indicate fewer unsupported unique-origin answers.}
\label{tab:rq3_p0_p3}
\centering
\footnotesize
\begin{tabular}{lccccc}
\hline
\textbf{Model} & \textbf{P0 FAR [95\% CI]} & \textbf{P3 FAR [95\% CI]} & \textbf{$\Delta$ FAR [95\% CI]} & \textbf{P0 UAR} & \textbf{P3 UAR} \\
\hline
GPT-5.5 & 87.5 [79.2, 94.4] & 38.9 [27.8, 50.0] & $-48.6$ [$-62.5$, $-34.7$] & 3.5 & 7.6 \\
GPT-5.6 Sol & 44.4 [33.3, 55.6] & 13.9 [6.9, 22.2] & $-30.6$ [$-43.1$, $-18.1$] & 24.3 & 34.0 \\
DeepSeek-V4-Pro & 100.0 [100.0, 100.0] & 100.0 [100.0, 100.0] & 0.0 [0.0, 0.0] & 0.0 & 0.0 \\
Claude Opus 5 & 12.5 [5.6, 20.8] & 0.0 [0.0, 0.0] & $-12.5$ [$-20.8$, $-5.6$] & 0.0 & 0.7 \\
Qwen 3.7 Max & 100.0 [100.0, 100.0] & 100.0 [100.0, 100.0] & 0.0 [0.0, 0.0] & 0.0 & 0.0 \\
\hline
\end{tabular}
\end{table*}

\subsubsection{Baseline Abstention.}
No model returns a valid unique-origin answer on a Sparse-Ambiguous row under any prompt, yielding 0\% FAR in all 20 model--prompt combinations. The P0 results differ sharply on Rich-Ambiguous rows. Claude Opus 5 has 12.5\% FAR, GPT-5.6 Sol has 44.4\%, GPT-5.5 has 87.5\%, and both DeepSeek-V4-Pro and Qwen 3.7 Max have 100\%. An explicit abstention rule therefore does not produce uniform behavior even though every ambiguous pair satisfies the same exact-payload equality criterion.

\subsubsection{Prompt Effects.}
P1 and P2 produce mixed model-dependent changes, while P3 yields the largest main-panel reductions for three models. Relative to P0, P3 lowers FAR by 48.6 points for GPT-5.5, 30.6 points for GPT-5.6 Sol, and 12.5 points for Claude Opus 5, with all three intervals excluding zero. DeepSeek-V4-Pro and Qwen 3.7 Max remain at 100\%. Under P3, unsupported answers concentrate on activation, symptom, and terminal events rather than the injected origin.

\subsubsection{Safety--Utility Trade-off.}
P3 also changes behavior on answerable inputs. Rich-Answerable UAR increases from 3.5\% to 7.6\% for GPT-5.5, from 24.3\% to 34.0\% for GPT-5.6 Sol, and from 0\% to 0.7\% for Claude Opus 5. DeepSeek-V4-Pro and Qwen 3.7 Max remain at 0\% UAR because they continue to return unique origins. Safe accuracy, which also requires the selected origin to be correct on answerable rows, rises from 74.1\% to 86.1\% for GPT-5.5 and from 93.4\% to 97.0\% for Claude Opus 5. It changes from 74.7\% to 75.7\% for GPT-5.6 Sol, falls from 38.0\% to 25.9\% for DeepSeek-V4-Pro, and falls from 38.9\% to 26.6\% for Qwen 3.7 Max. Lower FAR alone therefore does not determine the overall safety--utility outcome.

\subsubsection{Holdout and Repeatability.}
On the holdout, P3 again lowers FAR for four models and leaves Qwen 3.7 Max unchanged, while UAR increases for three models. The three-repeat study preserves this direction for the same four models in every repeat. Prompt benefits and abstention costs therefore remain model- and sample-dependent even when their direction is repeatable.

\textbf{RQ3 answer.} Models do not abstain uniformly when the visible payload is compatible with multiple origins. Contrastive elimination reduces unsupported unique-origin answers for GPT-5.5, GPT-5.6 Sol, and Claude Opus 5, while DeepSeek-V4-Pro and Qwen 3.7 Max remain unsafe under all four prompts. Holdout and repeated-call results show that prompt effectiveness is model- and sample-dependent and can raise unnecessary abstention.

\section{Conclusion}

TelemetrySuffBench separates three questions that are often collapsed in agent-system diagnosis: whether an execution failed, where the failure originated, and whether the available evidence supports any unique origin. Across five frontier models, the experiments show that these questions require different evidence and should be evaluated with different metrics.

For RQ1, Metadata, OpenTelemetry-compatible, and OpenInference-compatible views retain 99.5\%--100.0\% failure-detection F1 while limiting origin-step accuracy to at most 0.5\%. Detecting a downstream manifestation therefore provides little evidence about the event that introduced it. TraceElephant extends the Full-over-redacted localization pattern across five models and three public agent systems. For RQ2, removing decision content reduces origin-step accuracy to zero for every model. The matched evidence-gate pilot reproduces this collapse under a second controlled mechanism: Full localizes 92.0\% of fault origins and Full-D localizes none, while retaining 96.7\% detection accuracy. Provenance and identity make large positive contributions, while downstream state, verification, and terminal evidence cannot recover the origin on their own. For RQ3, contrastive elimination reduces the Rich-Ambiguous false-answer rate by 12.5--48.6 percentage points for three models in the main panel. Two models continue to return a unique origin on every Rich-Ambiguous case under all four prompts, and the holdout results show that prompt effects and abstention costs vary across models and samples.

Taken together, the results establish telemetry sufficiency as a task-relative property. A useful diagnostic trace must expose the decision that introduced the fault and preserve the identity and provenance links that connect that decision to a concrete event. An operational system must also represent when those links do not distinguish a unique origin. Reporting detection alone, or treating every complete-looking trace as attribution-ready, obscures both requirements.

The main benchmark uses one synthetic delayed-binding mechanism family, while the smaller evidence-gate pilot supplies a second matched mechanism under one fixed evaluator. Both satisfy the same causal-origin and exact-equivalence gates. TraceElephant provides RQ1 external evidence but lacks the factor crosswalk and matched origins required for RQ2. Most main-panel conditions use one model response per unique request, with repeated calls evaluated on a smaller matched subset. In RQ3, the Rich-Answerable and Rich-Ambiguous conditions are not matched on every payload-level distributional cue. These boundaries define the scope of the matched-ablation conclusion.

Future work will add matched interventions to production traces and evaluate whether explicit decision-to-object bindings and provenance edges improve localization.

\bibliographystyle{IEEEtranBST2/IEEEtran}
\bibliography{references}

\end{document}